\RequirePackage{fix-cm}
\documentclass[smallextended]{svjour3}       
\smartqed  
\usepackage{graphicx}
\usepackage{amsmath}
\journalname{Socio-Cognitive Systems}
\usepackage{algorithm}
\usepackage{algpseudocode}

\begin{document}

\title{Towards Social Identity in Socio-Cognitive Agents}

\author{Diogo Rato         \and
        Samuel Mascarenhas \and
        Rui Prada
}

\authorrunning{D. Rato, S. Mascarenhas, R. Prada} 

\institute{D. Rato
                \at INESC-ID \& Instituto Superior T\'ecnico, University of Lisbon, Portugal \\\email{diogo.rato@gaips.inesc-id.pt}
        \and 
            S. Mascarenhas
                \at INESC-ID, Lisbon, Portugal \\\email{samuel.mascarenhas@gaips.inesc-id.pt}
        \and
             R. Prada 
                \at INESC-ID \& Instituto Superior T\'ecnico, University of Lisbon, Portugal \\\email{rui.prada@gaips.inesc-id.pt}
}

\date{}

\maketitle
\begin{abstract}
Current architectures for social agents are designed around some specific units of social behaviour that address particular challenges. Although their performance might be adequate for controlled environments, deploying these agents in the wild is difficult. Moreover, the increasing demand for autonomous agents capable of living alongside humans calls for the design of more robust social agents that can cope with diverse social situations. We believe that to design such agents, their sociality and cognition should be conceived as one. This includes creating mechanisms for constructing social reality as an interpretation of the physical world with social meanings and selective deployment of cognitive resources adequate to the situation. We identify several design principles that should be considered while designing agent architectures for socio-cognitive systems. Taking these remarks into account, we propose a socio-cognitive agent model based on the concept of Cognitive Social Frames that allow the adaptation of an agent's cognition based on its interpretation of its surroundings, its Social Context. Our approach supports an agent's reasoning about other social actors and its relationship with them. Cognitive Social Frames can be built around social groups, and form the basis for social group dynamics mechanisms and construct of Social Identity.

\end{abstract}

\section{Introduction}
\label{sec:intro}
As social beings, people adapt their behaviour to live alongside other social actors that surround them. From simple actions, such as greeting another person, to more complex actions such as collaborating in group tasks, each individual's attitude changes to fit the reality that he/she is placed in. Nevertheless, the decision-making processes are not solely guided by external social pressures; people have their own preferences and follow their own motives as well, either influenced by their personality or past experiences. With multiple conflicting pressures in their daily life, people are recurrently faced with situations where they need to balance between choosing what is socially acceptable, pleasing others or their own desires. Furthermore, each person has their own considerations about what is socially acceptable. These multiple views of social conformity further increase the range of distinct and unique behaviour that people may exhibit. In addition, their uniqueness is not only present during the decision making process. In people's daily encounters with other social actors, the interpretation of their surroundings is a subjective process that enables each individual to construct their own reality. This construal process attaches a social dimension to the sensory perceptions that grant each individual the capability to view the shared world from a unique perspective~\cite{blumer1986symbolic}. All the above remarks highlight how social behaviour is deeply linked to cognition. Both influence each other to support the emergence of collective social phenomena as well as adapting to individuals motives.

With the increasing presence of autonomous machines in people's everyday life, more efforts to integrate them into current societies should be made. To achieve that, those machines must be endowed with social capabilities that allow them to understand people, adjust to their everyday life and to some extent, behave in a comparable manner. To become ubiquitous among humans, not only must they interact with the environment in a believable way but also reason over their actions' social aspects. To grant them the latter, they must comprehend the world as a social environment very much like humans do. Social notions, such as norms, categories, groups, values, or roles, must be embedded in the agent's cognition as they are also an integral part of a humans' cognition~\cite{clancey1997conceptual}.
Moreover, it is also hard to dissociate intelligence from social life since the evolution of both is profoundly linked~\cite{castelfranchi1998modelling}. By interacting with others, humans develop connections among them and, as demanded by each situation, people adapt their behaviour by including other social actors in their cognitive processes~\cite{clancey1991situated,smith2004socially}. Additionally, through their interpretation of the physical world, people mentally construct a unique representation of their social reality in their minds~\cite{castelfranchi1994guarantees}. Accordingly, there is a strong coupling between cognition and social behaviour, and this connection is bidirectional. Through social interactions, people are capable of improving and modifying their cognition while they deploy their cognitive resources as needed based on the social context they are placed in. 

Endowing agents with social capabilities should be a larger effort than just adding encapsulated social notions into agents, ironically similar to the way the word ``social'' was added to ``agents'' to create social agents. Attempting to shape particular modules of ``sociality'' into agents will not benefit their performance in real and complex social worlds~\cite{dignum2014autistic}. While recognizing the utility of designing specialized agents to address very precise tasks, applying an \textit{ad hoc} approach to the design of social agents will not promote the appearance of general intelligence. Similarly, the inclusion of units of social behaviour will prevent the conception of general social intelligence in agents. Looking forward to developing agents that can be included in people's everyday life, the methodology used to design them must introduce human's social behaviour in all of their components, from the interaction with other social actors to the selection of cognitive processes based on the situation. Therefore, to see an increasing successful presence of social agents in people's life, we argue that novel architectures for socio-cognitive agents should be researched and developed.

In the remaining sections, towards linking cognition to the social intelligence of agents, we elaborate on some concerns that should be taken into account by researchers. We enumerate five design principles that should be considered throughout the design of agents for Socio-Cognitive Systems. Then, we detail a computational model designed around the concept of Cognitive Social Frames inspired by Social Identity Theory with special emphasis on our notion of Social Context. Finally, we discuss some examples that can benefit from using our Cognitive Social Frames, and we conclude with some perspectives on future research on Socio-Cognitive Systems, namely, their importance for the integration of agents in our society.
\section{Design Principles}
\label{sec:design}

Despite the increasing presence of autonomous agents in people's daily lives, several researchers have highlighted the importance of understanding the social world in each individual's cognition. This connection and its implications have been a research topic that has called the attention of multiple theorists that study this phenomenon, and computer scientists that recognize its importance in the design of intelligent agents. In the late '90s, Castelfranchi argued that intelligence is a social phenomenon and that in Artificial Intelligence, researchers construct socially intelligent systems to understand it~\cite{castelfranchi1998modelling}. However, more recently, AI's goal regarding social intelligence has shifted towards designing and developing social agents that live alongside humans in a socially adequate manner. Nevertheless, the need to create social agents should not promote the inclusion of modules of social behaviour as additions to the agent's cognition~\cite{dignum2014autistic}. Instead, an agent's social nature should spread through its cognition.

To guide the design of socio-cognitive agents, we elaborate on five design principles that should be taken into account. For each one of them, we review the relevant background from social sciences and highlight some relevant works in Artificial Intelligence that explore it. 

\subsection{Socially Situated Cognition}
\label{ssec:sociallysituatedcognition}
Humans' actions are influenced by external factors. From characteristics of the environment that restrict physical capabilities, to the presence of others that demand an adaptation of the speech tone; for example, people adapt their behaviour to the situation they are placed in. Clancey proposed the notion of \textit{Situated Cognition} to explain the process people use to adapt their cognition based on their surroundings~\cite{clancey1991situated}. His proposal states that \textit{``all processes of behaving, including speech, problem-solving, and physical skills, are generated on the spot [...]''}. Primarily motivated to describe how knowledge is constructed in human's minds, the author suggested that people's cognitive resources are deployed as required by the context. In Clancey's proposal, knowledge is described as inherently social, similar to human actions, defining human's activity as socially oriented as well as socially shaped~\cite{clancey1997conceptual}. Similarly, Suchman also argued that not only physical but also social context can guide, restrict, or even fully decide people's actions~\cite{suchman1987plans}.

To study situated cognition in a social context, Smith and Semin introduced the concept of Socially Situated Cognition~\cite{smith2004socially}. It proposes four different properties of cognition: cognition is for action, cognition is embodied, cognition is situated, and cognition is distributed. The authors extensively reviewed evidence that situated action enables the appearance of social cognition. For instance, Schwarz argues that the perceivers' perspective of the context strengthens the sensitivity to external artefacts during attitude construal~\cite{schwarz2007attitude}.

We consider Socially Situated Cognition as one of the most important design principles to endow social agents with the capability to selectively deploy their cognitive resources as the social environment requires. Although generating cognitive resources on the spot is not suitable for computational models, as Clancey proposed, we agree that its capabilities should evolve with the agent's experience. Moreover, the sensitivity to the social context promotes the recruitment of only a part of the cognition that is deemed necessary, reducing the computational load on complex systems.


Socio-cognitive agents must be able to understand the situation (e.g., the adequate social roles and norms, and the relationship of the social actors) and act accordingly, but without losing the agency to choose whether to adhere to social pressures that the situation upholds.  In contrast to normative systems that restrict agents' actions to avoid a chaotic social world, we claim that socio-cognitive agents should not be enforced to blindly change their behaviour by an outside source. The agent's cognition should be the one promoting the balance between external pressures, such as social norms, and personal preferences. Overall, the principle can be summarized as:

\begin{center}
    \textit{A Socio-Cognitive Agent must selectively deploy its cognitive resources, hence adapting its behaviour, according to the Social Context.}
\end{center}

\subsection{Social Context and Construal}
\label{ssec:socialcontextandconstrual}
From the objects available around people to their relationships with other social actors, the entities that surround people have a direct influence on their cognitive processes. Nevertheless, not all humans ascribe similar meanings to the same entities they see in the physical world. In addition to the sensory experience that generates a perception, people construct mental representations of the world that are subject to an interpretative process.

One of the theories that explained the relationship of the self with the social reality is Symbolic Interactionism~\cite{blumer1986symbolic}. First proposed by George Mead and later published by Herbert Blume, the theory proposes that individuals interact with each other based on the symbolic world they construct. Since \textit{``humans act toward things on the basis of the meanings they ascribe to those things.''}, people can establish a link between their personal and unique view of entities and use it to adapt their cognition. Moreover, by arguing that \textit{``the meanings are handled in, and modified through, an interpretative process used by the person in dealing with the things he/she encounters''} the authors emphasize that interpretation plays an important role when trying to understand the social world people interact with. 

When endowing intelligent agents with the ability to selectively filter their computational processes based on the context they are placed in, there needs to be an explicit connection between the external world and their internal resources. Castelfranchi proposed that both physical context (relation to the environment) and social context (relation to other social actors) were noteworthy elements that influenced the agent's autonomy~\cite{castelfranchi1994guarantees}. Additionally, several researchers identified the importance of social context in the classification of agents as ``socially believable''~\cite{dautenhahn1998art,fong2003survey}. To be portrayed as such, an external observer should be able to perceive as meaningful the connection between the agent's identity, its behaviour and the social context.

From embodied virtual agents to social robots, the inclusion of autonomous agents in people's society requires them to understand people's current practices and adhere to them when appropriate. Using the physical reality might enable them to understand what the environment affords. However, we argue that this is not enough. In addition to exploring the deployment of cognitive resources based on world entities with physical properties, agents must rely on the relationship between such elements and their frame of reference. Endowing them with the capability to reason on top of the social context that represents a network of relationships between objects and social actors will improve their chances of being perceived by humans as socially believable. Following some theorists' proposals, we argue that the design of social agents must include a representation of the Social Context, that results from an interpretative process that refines and extends the sensory stimulus with social meaning. Therefore:

\begin{center}
    \textit{A Socio-Cognitive Agent must be able to construct Social Context by means of interpretative processes that ascribe social meaning to its sensory information.}
\end{center}

\subsection{Social Categorization and Identity}
\label{ssec:socialidentity}

Allport studied human's natural process for thinking about others in terms of their group memberships by means of Social Categorization~\cite{allport1954nature}. As part of their social nature, people tend to categorize other social actors on the basis of their social groups' memberships. Therefore, being able to recognize others' social groups, a person can also identify himself based on his relationship to such groups. This interaction between one's identity as a group member and the inter-group's dynamics was elaborated by Social Identity Theory~\cite{tajfel1981human}. It proposes that people are capable of constructing their social identity based on their relationship with other social groups. Later, Adams and Hoggs predicted that social identities can be taken as required by the situation, the Social Context~\cite{abrams1999social}. Together, both formulations suggest that humans can assume an identity as they see fit while being capable of recognizing, and comparing to others' memberships. Brewer further elaborated on the inter-group behaviour, by proposing that one, when adopting a social identity, balances the similarities with other in-group members and the distinction between in-group and out-group~\cite{brewer2001ingroup} members.

Owens et al.~\cite{owens2010three} studied two types of identity: \textit{role identities} and \textit{personal identities}. The first refers to identities that can only exist when a relevant counter-part is also present, for example, one cannot be a mother without children. However, personal identity reflects one's traits or characteristics that are not linked to others' social or role identities. As such, role identities present an interesting way of adjusting one's cognition based on the relationship with the reality, namely, other social actors.

The application of Social Identity in computational systems is also a challenging task. To design social agents, the notion of social context cannot be the only factor considered. If so, the individual nature of cognition can be lost. Therefore, the inclusion of aspects related to Social Identity can stimulate the individual differences of agents~\cite{clancey1997conceptual}. This distinction can be traced to people's different emotional attachment to each social group. Therefore, besides the interpretation of the surroundings, socio-cognitive agents' architectures must also include mechanisms that allow them to reason about other actors' memberships and how one fits into the social world.

Social Identity Theory can provide insight into the mechanisms an agent must accommodate to identify social categories and be able to define social groups. Furthermore, when placed together with other social actors, social agents must be capable of understanding the relationship between social groups. Ultimately, while endowed with the possibility to recognizing others' memberships and relate itself to them, an agent should be capable of constructing its own social identity that reflects its preferences and personal history in the social world. Similar to how Social Context represents the agent's interpretation of the physical reality, its Social Identity serves as a manner to identify how an agent positions itself within the social landscape filled with other social actors distributed in social groups, so:

\begin{center}
    \textit{A Socio-Cognitive Agent must be able to attribute social categories to social actors and understand their social meanings and relationship as part of the construction of Social Context. Additionally, it should be able to position itself in the social categorization space and have preferences over some social categories.} 
\end{center}

\subsection{Social Affordances}
\label{ssec:socialaffordances}

According to Gibson, affordances are the interactive opportunities offered by the environment to an organism~\cite{gibson1977affordances}. For instance, a book can afford several types of interactions ranging from interactions that are more common, such as, opening the book or reading its paragraphs, to other less orthodox interactions, such as using it as a cup holder or a wobbly table's stabilizer. However, an affordance does not live inside the organism nor the environment; it emerges from the ecological relationship between both parties~\cite{gibson1979ecological}. Moreover, affordances result from the coupling of perceptual information with the organism's cognitive capabilities. Since Gibson's theory of affordances was mainly conceived for direct visual perception, his proposal did not detail the role of cognition on the conception of affordances. Nevertheless, Gibson briefly challenged the affordances' original domain, alluding that other biological perceptions or cultural processes may allow the emergence of other types of affordances. In the past four decades, following his initial contributions, other researchers extended the original theory and explored the cognition's implications in recognition of affordances.

Some researchers proposed that perceptions result from a mental reconstruction of the physical world that is used to update our own internal representation of our surroundings~\cite{riesenhuber2002neural}. However, such theories did not consider the role of the organism's cognition in the perception process. By exploring the cognition's impact on the theory of affordances, other researchers suggest that perception through sensory inputs is highly influenced by the cognitive capabilities of an organism~\cite{cisek2010neural}. For example, Hirsh et al. proposed that perceptions can be understood as interpretations of the sensory input based on past experiences, expectations, and motives~\cite{hirsh2012psychological}. Others suggested that the perceiver's culture affects salience of some affordances, namely, the ones called social affordances~\cite{kaufmann2007culture}. Zhang and Patel, based on the distributed cognition framework, defined an affordance as a representation shared between the environment and the organism that can be categorized as \textit{biological}, \textit{physical}, \textit{perceptual}, or \textit{cognitive}~\cite{zhang2006distributed}, where the latter category refers to affordances provided by cultural conventions. Authors from distinct disciplines propose new interpretations to the ambiguous concept of affordance proposed by Gibson, but a common aspect was shared between their theories: cognition guides the attention mechanism that enables organisms to construct their internal representation of the physical reality. 

The theory of affordance is also relevant while designing computational models for agents, primarily to help construct perceptions based on the resources agents can deploy. Particularly, agents' architectures should consider potential interactions with other social actors as social affordances. Indeed, some computer science researchers have explored agents' social affordances across multiple domains. In accordance with Gibson's theory, Kreijns proposed a definition of social affordance to be applied in Computer-Supported Collaborative Learning~\cite{kreijns2001cscl}. The authors highlight the need for the CSCL environment to stimulate the group members' intervention while reciprocally when a member becomes salient to another, the social affordances must help guide the second member to engage in appropriate interaction with the first. The ecological stance of social affordances also found relevancy across the field of Human-Robot Interaction~\cite{min2016affordanceonrobots,wiltshire2017enabling}. Whether to improve a robot's planning capabilities by learning others' social affordances from the physical environment~\cite{pandey2012visuo,uyanik2013learning}, or by learning an affordance grammar from videos~\cite{shu2017learning}, approaches that use the theory of affordances can be found in multiple research works about social robots. 

In practice, we argue that social affordances, as proposed by Gibson and later reiterated by other researchers, present an interesting method for designing social agents, particularly to define their interactions with other social actors. By first identifying the bidirectional relationship between the perceiver and the surrounding social world, an agent can recognize opportunities for interaction with others. This principle can be summarized as: 

\begin{center}
    \textit{A Socio-Cognitive Agent must be able to identify social affordances in the presence of other social actors.}
\end{center}

\subsection{Socially Affordable}
\label{ssec:sociallyaffordable}

Alongside its ecological motivation, social affordances can also bring technical improvements for social agents. Gibson argued that perception is not about passively constructing an internal representation of the world, but rather about actively picking up information of interest to one’s behaviour~\cite{gibson1979ecological}. As such, an agent that only perceives information that is worth considering reduces its set of relevant perceptions. Computationally, it means creating an attention process that allows an agent to filter the perceptions based on their relevancy to its cognition. Nevertheless, this approach may not be sufficient to create agents that are more socially capable. Being able to recognize social affordances, opportunities for interaction with other social actors in the environment, does not necessarily mean that those potential interactions are adequate for the Social Context. Instead, understanding the distinction between what is possible, the mentioned affordances, and what is appropriate, hence, socially affordable, endows agents with the capability to adhere to social conformity. 

Let us revisit our previous example, the book's affordances are directly related to its capabilities and the physical environment the book is placed in. However, as social actors, people also take into account the setting of the interaction and what is deemed appropriate for each context. While using a book as a cup holder at home might not be questionable, the same action in a library might be inadmissible. The same principle applies to other types of affordances, such as social affordances. Whereas engaging a fellow reader in a library to dance can be considered inappropriate, the same interaction at home may be seen as adequate. To mimic this social awareness into agents through affordances, the environment must be perceived as an enabler of some opportunities that are socially affordable. Thus, an agent must be capable of representing and deploying social information regarded as relevant during the attention process, such as the relationship between other social actors and other physical entities. 

Such a distinction between affordances, including social, and what is socially affordable, emphasizes the separation between the actions that are deemed possible and the ones that are socially acceptable. This knowledge regarding social conformity grants social agents the capability to engage in interactions with the environment and other social actors that fit their context, therefore:

\begin{center}
    \textit{A Socio-Cognitive Agent must be able to recognize which affordances are socially affordable - adequate for the Social Context.}
\end{center}

\section{Cognitive Social Frames}
\label{sec:csf}
Humans live alongside other social actors. As members of society, we need to adapt our actions to fit others' expectations. Similarly, social agents must change their behaviour according to their reality. To endow this capability to socio-cognitive agents, from virtual agents to robots, we must develop mechanisms to adapt their cognition based on their interpretation of the physical reality. Taking into account the principles elaborated in Section ~\ref{sec:design}, we propose a computational model based on the concept of Cognitive Social Frame.

A Cognitive Social Frame (CSF) is the core element of a framework that enables the adjustment of the agent's cognition based on its interpretation of the surroundings. The latter is an internal representation of the agent's relationship with the things and other social actors placed in the world. This mental representation is called Social Context. The agent's cognition is encapsulated into several abstract blocks called Cognitive Resources. They contain specific knowledge and mechanisms used to determine the agent's actions. Finally, each CSF establishes a link between the Social Context and the relevant Cognitive Resources, that will be used to determine which Cognitive Resources should be deployed. In addition to the internal representation of these concepts, we also describe the agent's mechanism used to regulate the Cognitive Social Frames. We propose an approach with five stages that includes the process to interpret the perceptions and update the deployed Cognitive Resources.

In the remaining section, we detail the elements of the agent's architecture and how they are grouped. Then, we describe the several stages of the agent's mechanism and how its multiple components of the model interact with each other. Finally, we elaborate on some interesting phenomena that could be modelled with Cognitive Social Frames and how we align the model's goals with the design principles previously elaborated on.
\subsection{Architecture}
\label{ssec:architecture}

\begin{figure}[h]
\centering
\resizebox{0.9\hsize}{!}{\includegraphics{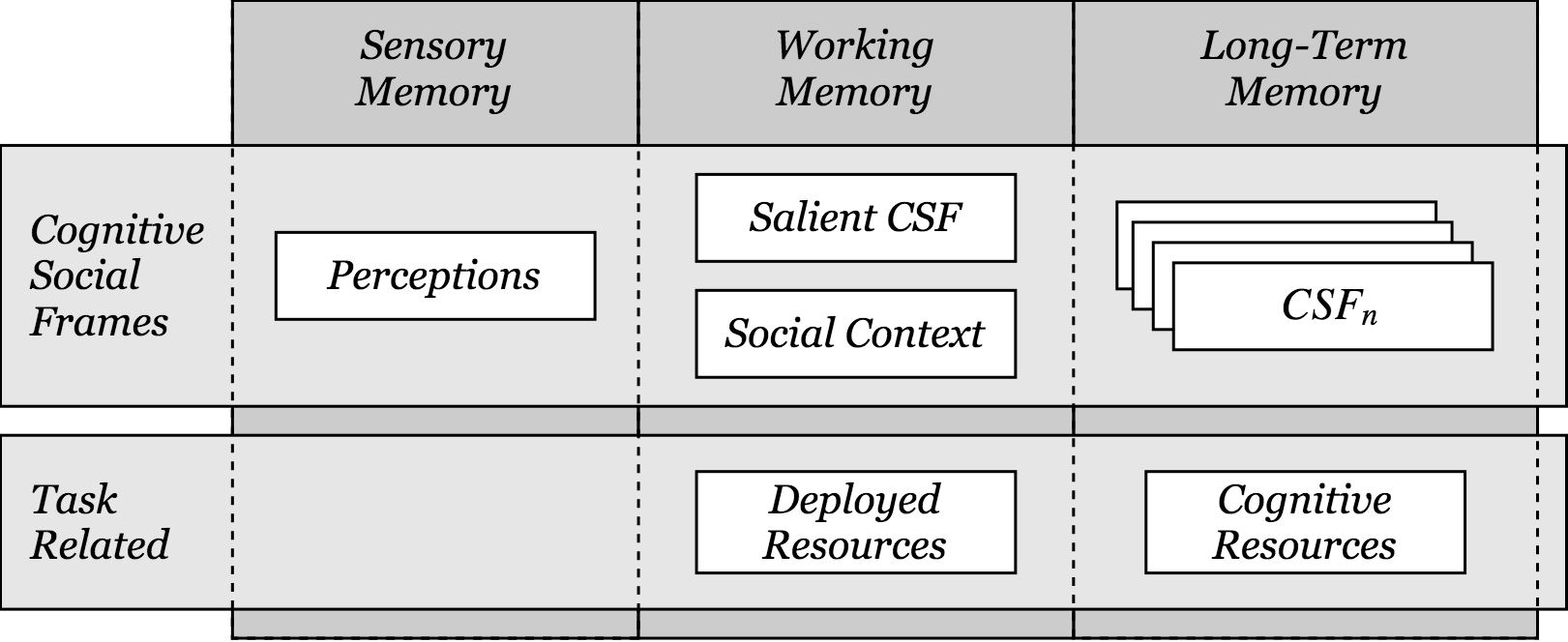}}
\caption{Cognitive Social Frame's architecture. Its elements are grouped across two dimensions: the relevancy for the task or the core CSF mechanism, and the memory type}
\label{fig:csfarchitecture}
\end{figure}

We propose an abstract architecture that supports the selective deployment of multiple cognitive resources depending on the agent's social context. To distinguish between the resources that are related to the core mechanism of the Cognitive Social Frames and the resources related to the task, we split them across two groups, as shown in Figure~\ref{fig:csfarchitecture}. Additionally, we also group the concepts in two memory groups that help manage the agents' resources: Sensory Memory, Working Memory, and Long-Term Memory.

Throughout its interactions with the environment, the agent will receive new perceptions and interpret them. This process results in new Social Contexts that will guide the adjustment of the agent's cognition. The connection between the Social Context and the agent's cognition is represented by the concept of Cognitive Social Frame. Although the agent holds a collection of possible CSFs, at a certain moment, only a subset of those is considered appropriate for the Social Context, called salient Cognitive Social Frames. Each CSF identifies the proper Cognitive Resources to deploy when it is salient. Our architecture does not impose limitations regarding the internal processes of each Cognitive Resource. However, some restrictions regarding the knowledge accessibility are established. 

\subsubsection{Memory}
The computational model's architecture organizes its elements into three types of memory: the Sensory Memory, the Working Memory, and the Long-Term Memory. Each memory type clusters concepts that have different lifetimes and salience.

\paragraph{Sensory Memory}
This memory group contains information about the agent's perceptions gathered from its sensors. The lifetime of the elements in this memory is very short as they are discarded right after they are interpreted by the agent. The Cognitive Resources of the agent, namely, the ones related to the tasks, do not use the sensory memory's perceptions directly. Instead, the agent's internal Cognitive Resources access its interpretation of the social world, the Social Context, rather than the unfiltered perceptions containing information about its physical surroundings.

\paragraph{Working Memory}
The Working Memory holds information about the current Social Context, the salient Cognitive Social Frames and other relevant data for the Cognitive Resources currently deployed. The elements of this memory group have a short lifetime as they are constantly being manipulated by other Cognitive Resources as well as by the core updating process of the Cognitive Social Frames. The Social Context is frequently updated by the agent based on the interpretation of new perceptions. This Working Memory's resource is used by the Cognitive Resources of the agents as well as by the salient Cognitive Social Frame updating mechanism. Alongside these two elements, the working memory also holds other memory resources used and managed by the Cognitive Resources deployed.

\paragraph{Long-Term Memory}
The elements inside this memory group do not change as often as the ones from the Sensory and Working Memory. In addition to storing other persistent elements used by the reasoning resources, the Long-Term Memory holds information about all the Cognitive Social Frames. Although not repeatedly modified, these elements are deployed as requested by the agent's other processes, such as the CSF's update mechanism.

\subsubsection{Cognitive Social Frame}

A Cognitive Social Frame is a representation that describes the relationship between the agent's Social Context and its available Cognitive Resources. At a certain instant, multiple CSFs can be appropriate for the same Social Context. Similar to the terminology used by the Social Identity Theory, we use the term \textit{salience} to describe the current relevancy of each CSF. In addition to that, each CSF defines a construal mechanism that is triggered when the agent perceives the environment. It is a process that supports the creation of a Social Context as a personal interpretation of the physical world rather than just a collection of sensory stimuli. A Cognitive Social Frame can be formally defined as
\begin{displaymath}
CSF = \{construal, fitness, CognitiveResources\},
\end{displaymath}

\paragraph{Construal} 
is a function that, based on the agent's perceptions in the Sensory Memory, constructs a Social Context. This function can be seen as a two-phase process. The first, inspired by Gibson's Theory of Affordances, resembles an attention mechanism that filters the agent's perception based on its relevancy for the respective CSF. The second phase, motivated by the interpretative view of a human's perception as proposed by Symbolic Interactionism and Social Identity Theory, transforms the set of perceptions into a collection of social perceptions that carries the agent's relationship to things and other social actors. The two phases combined result is a function that creates a Social Context that will guide the agent's deployed cognitive resources. The construal function signature is: 
\begin{displaymath}
Construal: Perceptions \longmapsto SocialContext
\end{displaymath}

\paragraph{Fitness} is a function that tests if the Cognitive Social Frame fits the agent's context. Both the external pressures, defined in the Social Context, and  the internal drives of the agent, present in the Cognitive Resources inside Working Memory, should influence the salience of the CSF. The outcome of the fitness function will contribute to determining whether or not the CSF should be considered salient.

\begin{displaymath}
Fitness: Working Memory \longmapsto ]0, 1]
\end{displaymath}

\paragraph{Cognitive Resources}
is the collection of the Cognitive Resources that will be deployed when the Cognitive Social Frame is considered salient.

\subsubsection{Social Context}

The agent's Social Context is a mental construct that describes the agent's interpretation of the surroundings at a given time. Without restricting the structure of its content, our computational model suggests that the underlying representation of the Social Context should be similar to the one used to represent perceptions. Conceptually, the units of knowledge that compose the perceptions directly gathered from the environment should be much alike than the ones that constitute the Social Context. However, in practice, the ones that form the latter must carry a social dimension that the former does not support, resulting in Social Perceptions.

The genesis of a Social Context is the result of a two-phase construal process. The first focuses on the agent's attention, filtering the perceptions, whereas the second annexes a social layer that results from an interpretation of such perceptions. As such, Social Context can be defined as follows
\begin{displaymath}
Social Context = \{SocialPerception_1, ...SocialPerception_n\}.
\end{displaymath}

\subsubsection{Cognitive Resources}

Our model does not focus on providing mechanisms to address particular challenges, such as planning, decision-making, and natural language processing. Nonetheless, the abstract nature of our framework includes room for such resources to be included. Within our architecture, these elements are encased in Cognitive Resources. For instance, a cognitive resource can be used to represent \textit{knowledge}, \textit{drives} and \textit{restrictions} for behaviour, such as a set of beliefs, motivations, or norms, or can represent \textit{mechanisms} such as decision-making processes or planning techniques. From the Cognitive Social Frames' perspective, each Cognitive Resources is an independent service that is used to generate behaviour for specific tasks.

\begin{figure}[h]
\centering
\resizebox{0.8\hsize}{!}{\includegraphics{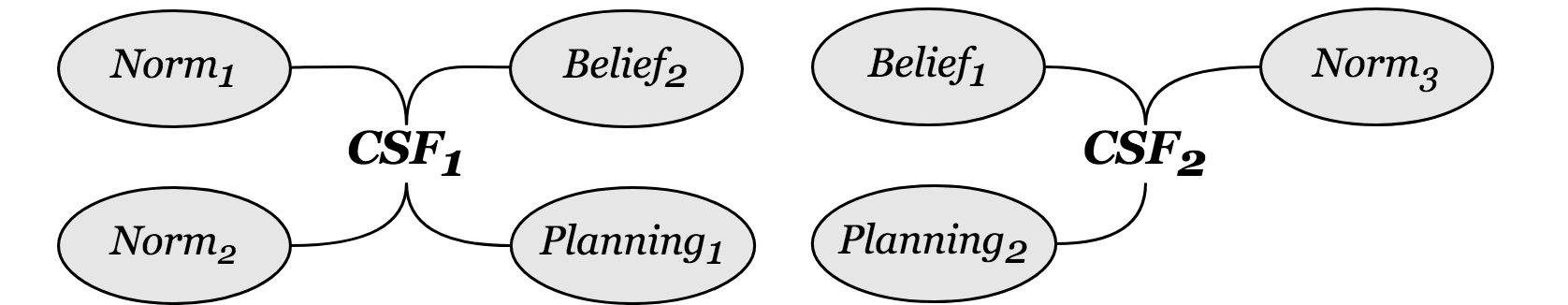}}
\caption{An abstract representation of several Cognitive Resources associated with two different Cognitive Social Frames, $CSF_1$ and $CSF_2$. Three types of Cognitive Resources are depicted, namely, to represent norms, beliefs and a planning mechanism.}
\label{fig:cognitiveresources}
\end{figure}

Although our model is not directly concerned with the contents of each Cognitive Resource, the agent is responsible for deploying them as befitting for the current state of the agent, that is, as prescribed by the salient Cognitive Social Frames. However, due to the high abstraction drawn between the deployment of Cognitive Resources and their internal processes, a Cognitive Resource may compromise the boundaries of other deployed resources by freely accessing everything in the agent's memory. Therefore, we introduce some rules to preserve the resources deployed by each CSF; otherwise, a Cognitive Resource would be able to ignore the mechanism that sets the salient Cognitive Resources based on the Social Context. Therefore, Cognitive Resource information management should obey some restrictions regarding its accessibility:
\begin{enumerate}
    \item A Cognitive Resource does not have access to the Sensory Memory. As a result, all observations of the world should be made through the Social Context.
    \item A Cognitive Resource can use all the information available in the Working Memory whether or not it was maintained or produced by itself or other resources.
    \item A Cognitive Resource has access to the Cognitive Social Frames in Long-Term memory. Therefore, it can rely on knowledge about the representation of other CSFs, besides the salient ones. 
\end{enumerate}

\subsection{Agent's Mechanism}
An agent's capability to perceive the environment and produce actions is also how our model defines the agent's interaction with the environment. However, rather than following the classic perceive, think and act loop, we claim that some mandatory additional steps should be included for socio-cognitive agents. We suggest an approach that defines the agent's mechanism as a five-stage process: \textit{Perceive}, \textit{Interpret}, \textit{Update}, \textit{Execute} and \textit{Act}. Each one of these phases manipulates different concepts present in the agent's architecture. Algorithm~\ref{alg:cycle} highlights the core steps in the agent's cognition from the instant it perceives the environment to the moment it acts on the world.

\begin{algorithm}[h]
\caption{Agent's Mechanism}
    \begin{algorithmic}[1]
        \Function{Cycle}{$Environment$}
            \State $Perceptions = PERCEIVE(Environment)$ 
            
            \State $Social Context = INTERPRET(Perceptions, CSF_{salient})$

            \State $CSF_{salient},Resources_{deployed} = UPDATE(CSF, SocialContext)$
            
            \State $Actions = EXECUTE(SocialContext, Resources_{deployed})$ 
            
            \State $ACT(Actions)$
        \EndFunction
    \end{algorithmic}
    \label{alg:cycle}
\end{algorithm}

Following the abstraction previously stated between the Cognitive Social Frames core elements and the Cognitive Resources, the agent's mechanism does not cover details about the mechanisms inside each Cognitive Resource.

\subsubsection{Perceive}
The perception of the agent is the first phase of the agent's loop and represents the observation of the physical environment by the agent through sensory stimuli. The result is a set of perceptions that is transferred to the agent's Sensory Memory.

\subsubsection{Interpret}
This phase is responsible for filtering the perception in the Sensory Memory and constructing the Social Context of the agent. After filtering the Sensory Memory's perceptions based on the salient Cognitive Social Frame and then transforming them into other social perceptions based on the physical world, the resulting Social Context is transferred to the Working Memory. Algorithm~\ref{alg:interpret} enumerates the steps of this phase.

\begin{algorithm}[h]
\caption{Agent's Interpretation}
    \begin{algorithmic}[1]
        \Function{Interpret}{$Perceptions, CSF_{salient}$}
            \State $SocialContext = \emptyset $
            \For{\textbf{each} $csf \in CSF_{salient}$}
                \State $SocialContext = SocialContext \cup csf.Construal(Perceptions)$
            \EndFor
            \State \Return $SocialContext$
        \EndFunction
    \end{algorithmic}
    \label{alg:interpret}
\end{algorithm}

At the start of the interpretation phase, the Social Context is cleared. Then, the agent uses the construal process associated with each salient Cognitive Social Frame to transform the set of perceptions into a set of social perceptions, the Social Context. In spite of not being transferred from one execution of the agent's mechanism to the next, the Social Context still accommodates the impact of the agent's previous state when interpreting the perceptions. Since the interpretation phase considers the salient Cognitive Social Frames; the resulting Social Context accounts indirectly for the previous events that lead to the current set of CSFs. Instead of updating the previous Social Context, the proposed mechanism relies on other elements of the agent's memory that were previously updated to interpret the physical world. This level of indirection grants to the agent the capability to rapidly shift the Social Context when noteworthy perceptions are received.    

Because there are multiple salient CSFs at the same time, there will be several social contexts as well. Their union will result in a collection of social perceptions that can have ambiguity, that is, different CSFs can generate the same social perceptions. Furthermore, it can also produce conflicting social perceptions, since multiple CSFs can lead to distinct, eventually conflicting, interpretations of the same physical world. 

For instance, let us consider a person that is responsible for coaching the school's football team. Additionally, one of the kids on that team is this person's son. On the one hand, as a football coach, this person wants the team to perform well and, to do so, he needs to identify the problems that could impact the team's success. On the other hand, as a father, he wants to see his son play. During a game where his son is performing poorly, the father, based on the mentioned different views, interprets the situation differently. As a football coach he sees his son as responsible for the team's problems whereas as a father, he sees his son as deserving the opportunity to play. Modelling the father and coach roles as Cognitive Social Frames, when faced with the perception of his son, the two CSF construals produce conflicting social perceptions that coexist in the same Social Context: one that identifies the son as a liability, and another as a young player that deserves a chance.

The resulting Social Context can accommodate social perceptions that may lead to the deployment of conflicting goals or actions in the agent's cognition. This allows the emergence of internal dilemmas that should be addressed by the mechanisms encapsulated in the Cognitive Resources.

\subsubsection{Update}
After constructing the Social Context, the agent has a perspective on the surrounding reality. Therefore, the previous salient Cognitive Social Frames may not be appropriate anymore. This process selects the Cognitive Social Frames in the Long-Term Memory that satisfy the fitness condition. However, the salience is also determined by the personal preference of the agents. To consider the external pressure exerted by the Social Context and the agent's internal pressure, we propose that salience is calculated as a balance between fitness and preference. As a result, the Salience of a Cogntive Social Frame \textit{csf}, for the Social Context $SC$ can be defined as 
\begin{displaymath}
Salience: Cognitive Social Frame, Social Context \longmapsto [-1,1], 
\end{displaymath}
and should satisfy the following properties:
\begin{enumerate}
    \item The salience should be proportional to the value of Cognitive Social Frame fitness based on the Social Context, given by \textit{csf.fitness(SC)}. As such, the CSFs that are considered by the agent to be more appropriate than other CSFs should have a greater value of fitness.
    \item The salience should be proportional to the agent's preference for each Cognitive Social Frame, given by \textit{preference(csf)}. This property represents the agent's personal inclination towards each CSF.
\end{enumerate}

The second property of this function, mentions the \textit{preference} function. Conceptually, it represents the agent's choice from the set of available Cognitive Social Frames without considering the Social Context. That is, there might be an internal drive, need, or motivator, that influences the salience of a CSF. This function can be defined as: 
\begin{displaymath}
\textit{preference}: Cognitive Social Frame \longmapsto [-1,1]. 
\end{displaymath}

Overall, the Update stage is responsible for altering the agent's Salient CSFs in the Working Memory and, as such, modify the set of deployed Cognitive Resources. As shown in Algorithm~\ref{alg:update}, this phase uses the current $Social Context$ and the entire set of $Cognitive Social Frames$ present in the Long-Term Memory. The first step of this phase is to calculate the new set of Cognitive Social Frames. To do so, based on the current Social Context, the agent iterates through all the CSFs to determine if their salience is greater than a certain threshold $\epsilon_{salience}$. Then, based on the resources associated with all the members of the new set of salient CSFs, the agent identifies the new collection of cognitive resources that should be deployed. 

\begin{algorithm}[h]
\caption{Agent's CSF Update Mechanism}
    \begin{algorithmic}[1]
        \Function{Update}{$SocialContext,CSF$}
            
            \State $CSF_{salient} = \emptyset$
            \For{\textbf{each} $\textit{csf} \in CSF$}
                \If {$\textit{Salience(csf, SocialContext)} > \epsilon_{salience}$}
                    \State $CSF_{salient} = CSF_{salient} \cup \textit{csf}$
                \EndIf
            \EndFor
            
            \State $Resources_{deployed} = \emptyset$
            \For{\textbf{each} $\textit{csf} \in CSF_{salient}$}
                \State $Resources_{deployed} = Resources_{deployed} \cup \textit{csf.resources}$
            \EndFor
            \State \Return $CSF_{salient},Resources_{deployed}$
        \EndFunction
    \end{algorithmic}
    \label{alg:update}
\end{algorithm}

In our current model, the process of deploying new Cognitive Resources is a fairly simple substitution of the methods already present in the agent's Working Memory. However, this produces a sudden change in the set of deployed cognitive resources instead of a smooth transition where the previous resources would gradually lose relevance in the agent's cognition. Although discussing alternatives for implementing the deployment of new Cognitive Resources should depend on the scenario of the application of the agent; we propose two different approaches that could be followed. 

Rather than solely having a mechanism to deploy the Cognitive Resources, there could be a symmetrical process to \textit{undeploy} them. As such, after determining the new set of salient CSFs, the agent would calculate the subset of Cognitive Resources present in the Working Memory that need to be removed and trigger an undeployment process specific to each Cognitive Resource. Another possible approach is to apply the salience mechanism of the CSF directly to the Cognitive Resources. That is, each deployed Cognitive Resource would have a salience value that would continuously decrease with time, such as a decay. The
deployed Cognitive Resources would see their salience value renewed if the Social Context identified them, through the salient CSF resources, as worthy of being deployed and in this case maintained. When a Cognitive Resource' salience is lower than a certain threshold, it is then discarded from the set of deployed resources.

\subsubsection{Execute}
In this phase, the agent does not manipulate any core elements of the architecture directly associated with the Cognitive Social Frames model. Instead, the updated set of deployed Cognitive Resources, with access to the new Social Context present in the Working Memory, is executed. 

As previously described, the Cognitive Resources are the agent's units of cognition that target specific problem-solving, decision-making, reasoning capabilities, and others. They can either contain knowledge or processes. Only the processes have an executable nature and, therefore, they will be executed. Nevertheless, although not running any procedures, the Cognitive Resources that hold knowledge can also be manipulated by the other type of Cognitive Resources.

This phase can be characterized by the high degree of abstraction that encapsulates the agent's ability to reason, decide, or interact with the world, away from the mechanism used to guarantee its conformity with the social context. While executing the deployed Cognitive Resources, whatever the outcome, it is already in accordance with the agent's Social Context. Nevertheless, there are some Cognitive Resources that can use the core Cognitive Social Frames concepts, namely, the CSFs.

\subsection{Discussion}

The proposed computational model's goal is to enable the creation of socio-cognitive agents that have the capability to adapt their cognition according to the social context, i.e., their interpretation of the social world. Our model fulfils this goal, by introducing a mechanism based on the concept of Cognitive Social Frame, which work as the link between the agent's social context and its cognition. The design of the mechanism we propose was guided by the design principles previously described.

Aligned with the design principle of \textit{Socially Situated Cognition} (Section~\ref{ssec:sociallysituatedcognition}), the main motivation of Cognitive Social Frames is to establish a link between the agent's situation, formally represented in the Social Context and its cognition, encapsulated in its Cognitive Resources. As such, the agent is sensitive to the view of the surrounding world when deploying its cognitive resources. However, it is important to note that this sensitivity to the Social Context should not be confused with dependency. Although the agent takes into account the Social Context, using our model, its cognition's deployment is also influenced by its own preferences. 

A socio-cognitive agent implementing our mechanism can interpret the world, as stated in the \textit{Social Context and Contrual} principle (Section~\ref{ssec:socialcontextandconstrual}), rather than just perceiving it. The second stage of our mechanism allows a socio-cognitive agent to construct a mental representation, the Social Context, describing its relationship with perceived elements. This Social Context is the set of social perceptions that result from the application of each salient Cognitive Social Frame construal function. This function is responsible for filtering the agent's perceptions and then applying a social layer on top of them. As such, the Social Context enables the observer to allow its cognition to reason about the meaning of the elements of the physical world instead of the elements by themselves. 

Additionally, in the interpretation phase, the agent can construct a Social Context. This process is also influenced by the salient Cognitive Social Frames. Therefore, the interpretation of the reality is performed from the agent's frame of reference with regard to its relevance to the agent's cognition. As stated in the \textit{Social Affordances} principle (Section~\ref{ssec:socialaffordances}), a socio-cognitive agent should perceive what is worth paying attention to and identify social interaction opportunities in the Social Context. Our proposal supports this suggestion since it only applies the construal function of the salient CSFs relevant for the cognition, to create the Social Context. The principle described in Section~\ref{ssec:sociallyaffordable} emphasizes that a socio-cognitive agent should recognize what is socially affordable. In line with this remark, a Cognitive Social Frame represents the Cognitive Resources that are associated with a particular Social Context and, to a certain degree, it also dictates how the agent can interact with the world. Moreover, while building the Social Context, CSFs are attributed to other social actors as well. This supports the identification of social affordances, but at the same time is a mechanism that enables certain mind-reading capabilities in the agent.

Finally, one of the most promising aspects of our proposal is related to the principle of \textit{Social Categorization and Identity}(Section~\ref{ssec:socialidentity}). The concept of Cognitive Social Frames supports the appearance of the concept of Social Identity by enabling a socio-cognitive agent to identify its and others' social categories. When placed in a world with other social actors, an agent capable of representing the concept of CSF can also assign to others their salient CSFs. Furthermore, it can also reason about its beliefs regarding others' salient CSF and their social categories. However, this principle also claims that not only should a socio-cognitive agent recognize others' group memberships but also be able to construct its own social identity based on its relationship with the social category, by defining personal preferences over some identities. Regarding the first, a Cognitive Social Frame can be used as the concept that enables the categorization of social actors and, therefore, defines groups of social actors that share similar CSFs. With interest to the second, the mechanism we propose allows the Cognitive Resources to reason about the concept of Cognitive Social Frames and project into others' salient CSFs, modelling others' categories. With this information, an agent can explore its relationship with other social actors, considering their memberships, towards defining its own Social Identity.

In addition, by considering the salient CSFs of other social actors with its own cognition, an agent is capable of reasoning about others' deployed Cognitive Resources and, therefore, acknowledge their beliefs, goals, mechanisms, and others. This \textit{mind-reading} capability can enhance the social dimension of such agents since they can now expect and predict others' actions based on their salient Cognitive Social Frames. Furthermore, this mind-reading capability can be extended from the recognition of what Cognitive Resources are deployed to how another social actor interprets the physical world, thus creating Social Contexts from other frames of reference. Combining the two, the Social Context and Cognitive Social Frames, a socio-cognitive agent has, to an extent, a mind-reading mechanism that allows it to understand the world from others' perspectives and potentially anticipate others' behaviours. 

The ability to mind-read other social actors can help an agent establish relationships with other social actors. Instead of looking at the environment as a mere collection of opportunities for interaction, focusing on the interpersonal relationship with others creates agents that are more socially capable. While interacting with other social actors, an agent has a better chance of successfully engaging with them if it is aware of their drives, beliefs, norms and other aspects that can be derived from their salient Cognitive Social Frames. With this information, when interacting with other social actors, a socio-cognitive agent can search for common grounds with its interlocutors, thus strengthening their interpersonal relationship. We can use CSFs to explore the discrete (based on categories) nature of social relationships that are often treated as a continuous variable in multi-agent systems. For example, we can define a CSF for friends and another for acquaintances and define in each the social norms that apply when the agent meets other actors that fit the CSF. 

Additionally, agents with mind-reading capabilities can use their knowledge about others' interpretation of the reality to manipulate the constructing of their Social Context, in particular, the identity (e.g., salient CSF) that others ascribe to the agent. Managing others' impressions about itself requires a socio-cognitive agent to reason about others' construction of the Social Context. Looking forward to inducing perceptions that will influence other's views about itself, an agent can either adjust the information exchanged with others or modify the environment such that their construction of the Social Context alters the other's interpretation of the social reality.

\section{Conclusion}
\label{sec:conclusion}

Our everyday life takes place in social worlds that have started to welcome new intelligent autonomous agents to live alongside humans. To properly adjust them to people's social lives, they must be endowed with social capabilities that allow them to adapt their interaction with the surrounding social actors. We claim that social agents would be more capable of successfully engaging others if they could change their cognition based on the social context they are placed in. For example, agents should know the role they play in a given context and should be able to change the role if the context changes.

In this document, we identified some design principles that should be taken into account when conceiving socio-cognitive agents to live alongside other social actors. We explored the implications of some theories proposed by sociologists and psycho-sociologists in computational models. Overall, we look to social sciences as an extremely helpful source of knowledge when engineering social behaviour.

We proposed the Cognitive Social Frames model that establishes a relationship between a personal construct of the physical world, the Social Context, and the agent's Cognitive Resources. Its mechanism allows the agent to interpret the shared reality based on its salient CSFs and deploy different Cognitive Resources to match the requirements of the environment. Bounding the agents' relationship with reality, the Social Context, with its internal processes, the Cognitive Elements, our model establishes a framework that promotes the deployment of socio-cognitive agents alongside other social actors living in an environment filled with social rules and opportunities for interaction.



\begin{acknowledgements}
This work was supported by national funds through Portuguese Funda\c{c}\~ao para a Ci\^encia e a Tecnologia (FCT: UID/CEC/50021/2019). Diogo Rato acknowledges his FCT grant (SFRH BD/131024/2017).
\end{acknowledgements}

\bibliographystyle{spmpsci}      
\bibliography{myBib}

%
%

\end{document}